\documentclass[letterpaper, 10 pt, conference]{ieeeconf}  
\usepackage{xcolor}
\usepackage{soul}

\usepackage{balance} 
\usepackage{float}
\usepackage{amsmath}
\usepackage{multirow}
\usepackage{algorithm}
\usepackage[noend]{algpseudocode}
\usepackage{graphicx}
\usepackage{booktabs}

\IEEEoverridecommandlockouts                              

\title{\LARGE \bf
An End-to-End Collaborative Learning Approach for Connected Autonomous Vehicles in Occluded Scenarios
}

\author{Leandro Parada$^{1}$, Hanlin Tian$^{1}$, Jose Escribano$^{1}$, Panagiotis Angeloudis$^{1}$
\thanks{$^{1}$Leandro Parada, Hanlin Tian, Jose Escribano and Panagiotis Angeloudis are with the Centre for Transport Studies, Department of Civil and Environmental            Engineering, Imperial College London, London SW7 2AZ, United Kingdom
       {\tt\small lap20@ic.ac.uk}}%
}

\begin{document}

\maketitle
\thispagestyle{empty}
\pagestyle{empty}

\begin{abstract}

Collaborative navigation becomes essential in situations of occluded scenarios in autonomous driving where independent driving policies are likely to lead to collisions. One promising approach to address this issue is through the use of Vehicle-to-Vehicle (V2V) networks that allow for the sharing of perception information with nearby agents, preventing catastrophic accidents. In this article, we propose a collaborative control method based on a V2V network for sharing compressed LiDAR features and employing Proximal Policy Optimisation to train safe and efficient navigation policies. Unlike previous approaches that rely on expert data (behaviour cloning), our proposed approach learns the multi-agent policies directly from experience in the occluded environment, while effectively meeting bandwidth limitations. The proposed method first prepossesses LiDAR point cloud data to obtain meaningful features through a convolutional neural network and then shares them with nearby CAVs to alert for potentially dangerous situations. To evaluate the proposed method, we developed an occluded intersection gym environment based on the CARLA autonomous driving simulator, allowing real-time data sharing among agents. Our experimental results demonstrate the consistent superiority of our collaborative control method over an independent reinforcement learning method and a cooperative early fusion method.

\end{abstract}

\section{INTRODUCTION}

Unsignalised intersections pose a challenge because they lack traffic signals or signs, leading to confusion and potentially dangerous situations for drivers, pedestrians, and cyclists. Moreover, vehicles may not have a direct line of sight to all other vehicles and pedestrians within the intersection zone, creating occluded scenarios. These challenges can be potentially overcome with the introduction of Autonomous Vehicles (AVs) because they have the ability to exchange valuable information in real time. Producing autonomous systems that can safely navigate through these scenarios is, therefore, essential for the rollout of AVs. However, the number of potential scenarios that can occur in an intersection is too vast to implement rule-based methods.

Deep Reinforcement Learning (DRL) has recently been shown to be a powerful method for safely navigating occluded intersections \cite{kamran_risk-aware_2020, isele_navigating_2018}. Its relevance to this domain stems from the opportunity to derive policies by observing thousands of occluded scenarios with different vehicle configurations and intersection geometries. Previous studies on DRL have demonstrated the shortcomings of rule-based methods, which are considered state-of-the-art in handling challenging cooperative scenarios. Moreover, DRL has proven to be superior to these in terms of safety and traffic efficiency. However, the methods previously presented \cite{kamran_risk-aware_2020, isele_navigating_2018} consider just a single-agent setting; they do not use any form of collaborative perception and thus do not exploit the full potential of Connected Autonomous Vehicles (CAVs). 

\begin{figure}[t!]
  \centering
  \includegraphics[scale=0.4]{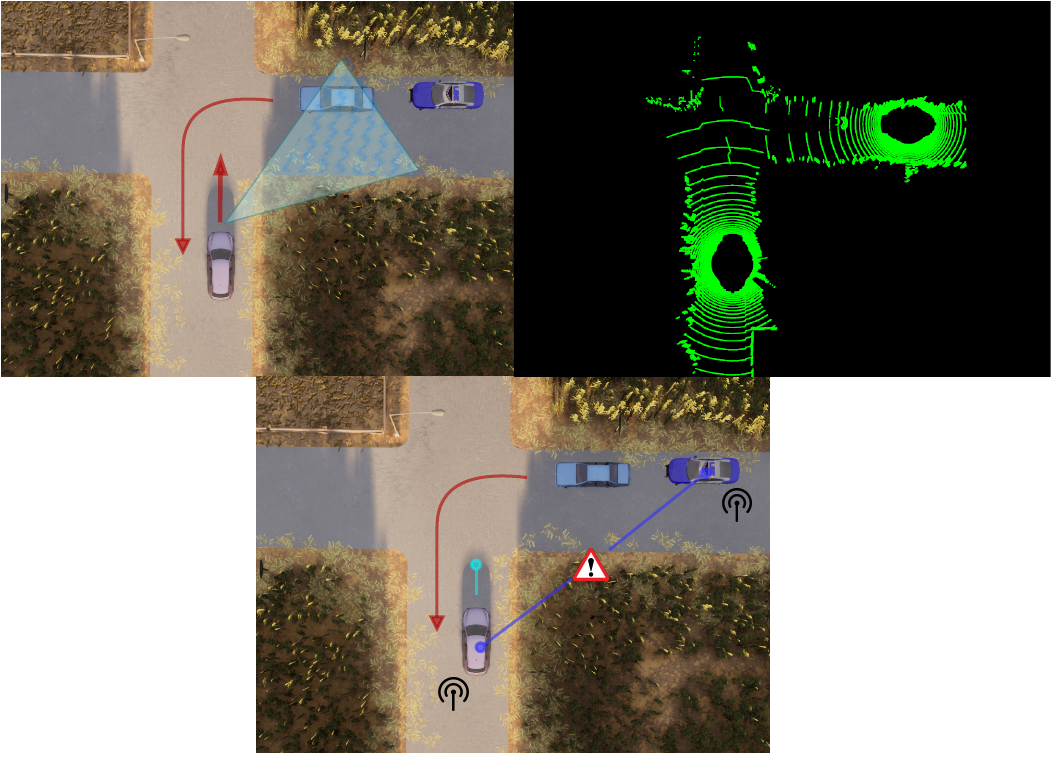}
  \caption{Collaborative CAV control. Some vehicles may be occluded by tall trees/buildings or other vehicles in the scene. With the use of LiDAR features from nearby agents, safe and efficient policies can be developed to avoid accidents.}
  \label{fig:summary}
\end{figure}

Collaborative perception is vital for Connected Autonomous Vehicles (CAVs) to accurately perceive the world in scenarios that involve visual occlusion. Consider a scenario such as the one presented in Figure \ref{fig:summary}, where a blocked vehicle in one lane becomes invisible to others, emphasising the need for collaborative perception to avoid a collision. Collaborative perception allows CAVs to exchange information via a vehicle-to-vehicle (V2V) ad-hoc network \cite{tonguz_broadcasting_2007}, ensuring real-time sharing of perception insights. However, sharing raw data in real-time poses computational and bandwidth challenges, prompting the need for efficient techniques aligned with hardware limitations.

To ensure secure and effective navigation in occluded environments, a collaborative effort from Connected Autonomous Vehicles (CAVs) becomes essential. Previous studies in this domain have focused on either collaborative Perception and Prediction (P\&P) \cite{xu_opv2v_2022, xu_v2x-vit_2022} or single-agent navigation \cite{isele_navigating_2018, kamran_risk-aware_2020, cui_coopernaut_2022}, but there has been limited work on collaborative navigation. The closest study to our approach is presented in \cite{isele_navigating_2018}, where the authors proposed a Deep Q-Network (DQN) to learn policies in occluded intersections. However, they modelled the problem from a single-agent perspective and did not consider collaborative navigation among CAVs.

A recent study \cite{cui_coopernaut_2022} addressed the problem of navigating occluded scenarios by proposing an end-to-end learning method that fuses LiDAR data from nearby vehicles and uses it to determine the control of a single vehicle (ego). In contrast to ours, their method is trained using expert data through behaviour cloning. Behaviour cloning suffers from various limitations \cite{codevilla_exploring_2019}, including bias of the driving dataset, causal confusion and high variance. Furthermore, acquiring expert data may not always be feasible or readily available, especially for certain edge cases.

In this study, we propose an end-to-end collaborative learning method to navigate dangerous scenarios involving occlusion. The method is based on a decentralised perception mechanism that preprocesses raw LiDAR data, which are then shared with nearby agents using an ad-hoc V2V network. The features extracted from LiDAR are received by nearby CAVs and utilised within a multi-agent reinforcement learning pipeline to obtain the desired driving action. Specifically, we use Multi-agent Proximal Policy Optimisation, a Multi-Agent Reinforcement Learning (MARL) method that has been shown to achieve superior performance in several cooperative games \cite{yu_surprising_2021}. In order to train our method, we design a gym environment based on the CARLA simulator that generates challenging occluded intersections with two or more CAVs and multiple human-driven vehicles.

The main contributions of this paper are:

\begin{enumerate}
\item We propose a collaborative control method that utilises LiDAR data to extract features and shares them with nearby Connected Autonomous Vehicles (CAVs) to navigate safely through challenging occluded scenarios.

\item Our approach utilises multi-agent reinforcement learning for training in occluded environments, distinguishing it from previous works that rely on expert data (behaviour cloning), which may not always be available in practice.

\item Our method surpasses the capabilities of Independent Reinforcement Learning (RL) and a cooperative early fusion method by providing safe and efficient policies, while also effectively addressing bandwidth limitations.

\end{enumerate}

\section{RELATED WORK}

\subsection{Intersection Management for CAVs}

Extensive research has been done on the navigation of CAVs in intersections. These include heuristics and rule-based methods and, more recently, DRL-based methods.

\subsubsection{Heuristics and Rule-based methods}

The majority of studies in the field of intersection management correspond to heuristics and rule-based methods. A study \cite{alonso_autonomous_2011} presented two rule-based priority methods for mixed scenarios, including manually driven cars. Later, another study \cite{zohdy_intersection_2012} proposed a Cooperative Autonomous Cruise Control system that optimises each vehicle's arrival time at the intersection stop-line. Their method reduced $91\%$ the average delay compared to a traditional intersection control method. In a different work, Hafner et al. \cite{hafner_cooperative_2013} proposed an efficient communication-based algorithm that creates a control throttle/brake map that keeps the vehicles from colliding with each other.

Other popular rule-based methods include reservation systems \cite{au_autonomous_2015, vasirani_market-inspired_2012, zhang_analysis_2013, wei_batch-light_2014} and Time-To-Collision (TTC) frameworks \cite{minderhoud_extended_2001, van_der_horst_time--collision_1994}. Reservation systems have been widely used and have achieved superior performance compared to other rule-based methods. These methods involve an intersection manager that simulates vehicle trajectories and assigns reservation slots to vehicles approaching the intersection. TTC has been used extensively as a safety indicator and has shown to be superior when compared to other rule-based methods \cite{minderhoud_extended_2001}.

Rule-based methods can be simple and easy to implement; thus, they are suitable for real-time intersection control. However, the complexity of these methods can substantially increase when adding more vehicles and constraints \cite{namazi_intelligent_2019}. A busy intersection may have several CAVs, human-driven cars and pedestrians. All of these entities can arrive at different times and velocities; consequently, the number of potential scenarios is enormous. TTC-based and other rule-based methods cannot possibly cover all scenarios that can occur in a busy intersection. Furthermore, occluded scenarios can be particularly challenging for traditional methods that do not rely on any form of collaboration. 

\subsubsection{Deep Reinforcement Learning}

The use of deep reinforcement learning (DRL) has shown promise in enhancing autonomous driving performance in challenging scenarios where traditional rule-based methods may be insufficient. For instance, Kamran et al. \cite{kamran_risk-aware_2020} proposed a DQN approach that penalizes risky situations resulting from occlusion. Their study demonstrated that the DQN risk-aware method could effectively learn in complex scenarios, while collision-based DQN and rule-based methods failed. Similarly, Isele et al. \cite{isele_navigating_2018} also used the DQN algorithm but relied on an occupancy grid to model the occluded area from the ego vehicle's perspective. They showed that their method could navigate successfully in scenarios where rule-based methods consistently failed. Additionally, Tram et al. \cite{tram_learning_2018} used metadata of nearby vehicles to predict vehicle intent and learn a policy that can negotiate, thereby demonstrating the potential of machine learning methods for improving autonomous driving performance in challenging scenarios.

Most of the DRL studies have approached the occluded navigation problem from a single-agent perspective \cite{isele_navigating_2018, kamran_risk-aware_2020, tram_learning_2018}. However, challenging occluded scenarios require active collaboration from CAVs in order to guarantee safe and efficient navigation. The integration of collaborative perception and decision-making mechanisms can enhance the capability of CAVs to share knowledge about the environment and take coordinated actions.

\subsection{Collaborative Perception}

Collaborative Vehicle-to-Vehicle (V2V) perception methods are categorized based on their data treatment strategy, which can be early fusion, intermediate or deep fusion, or late fusion. Early fusion methods communicate raw sensor data to neighboring vehicles, but this data can be difficult to handle in real-time due to its high dimensionality. Intermediate fusion methods fuse features extracted from raw data, while late fusion methods communicate perception results after the post-processing stage.

Recent studies in this field have explored intermediate fusion collaborative perception pipelines, which have shown promising results. For example, Xu et al. \cite{xu_v2x-vit_2022} presented a method that combines LiDAR data from nearby vehicles and applies a Vision Transformer for box regression and classification. Similarly, Vedaldi et al. \cite{vedaldi_v2vnet_2020} proposed a V2V communication method that aggregates LiDAR information into a compressed intermediate representation and achieves higher accuracy than other V2V communication baselines.

In addition, Xu et al. \cite{xu_opv2v_2022} compared different collaborative V2V perception methods over a new dataset called OpenV2V and found that intermediate fusion methods have a good balance between prediction accuracy and transmission bandwidth. Furthermore, Xu et al. \cite{xu_opv2v_2022} proposed a novel intermediate fusion method that uses a self-attention mechanism and outperforms all other fusion methods.

One of the main challenges of collaborative perception algorithms is algorithm scalability, particularly in settings with many agents. Various approaches have been proposed to address this challenge, including Who2Com \cite{liu_who2com_2020} and When2Com \cite{liu_when2com_2020}, which determine which agent to request/send data using a handshake mechanism and construct efficient communication groups to reduce the overall bandwidth load. Additionally, Aoki et al. \cite{aoki_cooperative_2020} proposed a method that uses Deep Reinforcement Learning (DRL) to automatically select which information to transmit to nearby CAVs to enhance sensing capabilities.

While most studies on autonomous collaborative systems focus on perception or perception and prediction \cite{xu_opv2v_2022, chen_cooper_2019, xu_v2x-vit_2022, liu_when2com_2020, liu_who2com_2020, rawashdeh_collaborative_2018}, few studies have addressed collaborative control. To fill this gap, we propose a multi-agent collaborative control method that extracts LiDAR point cloud features and shares them with nearby CAVs through an V2V network. This method is trained using model-free reinforcement learning and does not rely on a set of predefined rules.

\section{PRELIMINARIES}

\subsection{Multi-agent Proximal Policy Optimisation}

In the context of complex occluded scenarios, it is imperative to develop algorithms that are capable of generating cooperative decentralised policies. Oftentimes, a central controller may not be available or may prove to be inefficient, particularly when multiple vehicles are present in the scene. In such cases, it is crucial to implement solution algorithms that are distributed in nature. Multi-agent Proximal Policy Optimisation (MAPPO) is a state-of-the-art Actor-Critic algorithm that has been found to outperform other existing methods for complex cooperative games \cite{yu_surprising_2021}. MAPPO follows the Centralised Training and Decentralised Execution (CTDE) paradigm, which entails the use of a centralised state during the training phase for policy updates, and the decentralisation of actors during the execution phase.  

In MAPPO, the individual actors seek to maximise the following equation:

\begin{multline}
    L(\theta) = [\frac{1}{Bn} \sum_{i=1}^{B} \sum_{k=1}^{n}  min(r_{\theta, i}^{(k)}A_{i}^{(k)}, clip(r_{\theta, i}^{(k)}, 1-\epsilon, 1+ \epsilon) \\ A_{i}^{(k)}]  + \sigma \frac{1}{Bn} \sum_{i=1}^{B} \sum_{k=1}^{n} S [\pi_{\theta}(o_{i}^{(k)})],
\end{multline}

, where
\begin{equation}
    r_{\theta, i}^{(k)} = \frac{\pi_{\theta}(a_{i}^{(k)}| o_{i}^{(k)})}{\pi_{\theta_{old}}(a_{i}^{(k)}| o_{i}^{(k)})},
\end{equation}

where $A_{i}^{(k)}$ is calculated using Generalized Advantage Estimation (GAE) with advantage normalisation, $S$ is the policy entropy and $\sigma$ is the entropy coefficient. On the other hand, the critic minimises the following equation:

\begin{multline}
    L(\phi) = \frac{1}{Bn} \sum_{i=1}^{B} \sum_{k=1}^{n}(max[V_{\phi}(s_{i}^{(k)}) - \hat{R_{i}})^{2}, (clip(V_{\phi}(s_{i}^{(k)}), \\ V_{\phi_{old}}(s_{i}^{(k)}) -\epsilon, V_{\phi_{old}}(s_{i}^{(k)})+\epsilon)- \hat{R_{i}})^{2}],
\end{multline}

where $\hat{R_{i}}$ is the discounted error to go, $B$ is the batch size and $n$ the number of agents.

\section{METHODOLOGY}
\subsection{Occluded Intersection Gym Environment}

There are several autonomous driving gym environments available for evaluating DRL algorithms, such as SMARTS \cite{SMARTS}, Highway-env \cite{highway-env}, and MACAD \cite{palanisamy2019multiagent}. However, none of these environments satisfied our requirements for cooperative driving capabilities and realistic LiDAR point cloud representations. While some environments, such as SMARTS and MACAD, provide multi-agent capabilities, they do not have realistic LiDAR point cloud representations. On the other hand, Highway-env has realistic LiDAR point cloud representations but lacks cooperative driving capabilities. To address these limitations, we developed our own environment called the Occluded Intersection Gym Environment.

We developed our own gym environment using CARLA \cite{dupuis_opendrive_2010} and the OpenCDA framework \cite{xu_opencdaopen_2021} to simulate challenging cooperative scenarios in an episodic manner. The environment includes two or more CAVs and traffic vehicles in high-risk scenarios that may result in collisions. The CAVs can share information and take independent actions in real-time while traffic vehicles follow predetermined routes established by the SUMO simulator, without sharing information or taking any actions. We also introduced randomisation in specific simulation parameters, such as the types and characteristics of CAVs and traffic vehicles and their initial poses, so that we can train on different variations of the same scenario.

In each episode of our proposed gym environment, traffic vehicles follow one of several predefined routes through the intersection, all of which lead to challenging occluded scenarios. This presents a realistic and diverse set of scenarios for the CAVs to navigate safely and efficiently. To aid in this navigation, CAVs are able to share both metadata (such as pose and speed) and LiDAR features with nearby agents in real-time. We limit this sharing to agents within a predefined range, in order to maintain realism and avoid unrealistic global knowledge.

The elements of the gym environment can be summarised as follows: 

\begin{itemize}
    \item Agents: The agents correspond to the CAVs, which have partial observability of the environment and can share real-time information with other agents.
    \item Observation: Includes metadata from the ego vehicle and vehicles within a specific range. This data consists of the current speed, vehicle dimensions and current lane id. The raw LiDAR data consists of 4D points, where the first three dimensions correspond to the coordinates of each point $(x, y, z)$ and the last dimension represents the intensity loss during travel.
    \item Action: The action space consists of five discrete values; two for speeding up (smaller and larger amount), two for slowing down (smaller and large amount) and one for doing nothing.
    \item Reward: We assume a fully cooperative game; thus, agents share the same reward, defined by the following equation:
    \begin{equation}
        R = -5*r_{col} + r_{speed} + 5*r_{dest} + -0.01*r_{step}
    \end{equation}
    
   Where $r_{col}$ is a collision penalty, $r_{speed}$ is a positive reward for the speed of the vehicle, $r_{dest}$ is the reward given when the CAV reaches its desired destination and $r_{step}$ is a minor penalty given to encourage the agent to reach the goal promptly.
\end{itemize}

\subsection{Communication Message}

The V2V network facilitates communication among vehicles using compressed feature messages that encapsulate scene information from each CAV's perspective. We denote $M_i$ as the message shared by CAV $i$, given by $M_i = CNN(\hat{z}_i^{(0)})$, where $CNN$ comprises a sequence of convolutional layers and $\hat{z}_i$ represents the preprocessed LiDAR point cloud data. 

After testing various configurations, we found that utilising two convolutional layers yielded the best results while adhering to bandwidth limitations, which will be discussed in the Experiments section. The resulting feature size of the output is (4, 128, 128). It's important to acknowledge that our current message format is a custom design focusing on maximising message information while complying with bandwidth limitations. Future efforts should focus on developing a message format that aligns with established standards, such as the ETSI standard \cite{etsi_intelligent_nodate}.

\subsection{Collaborative MARL framework}

\begin{figure*}[t]
  \centering
  \includegraphics[scale=0.6]{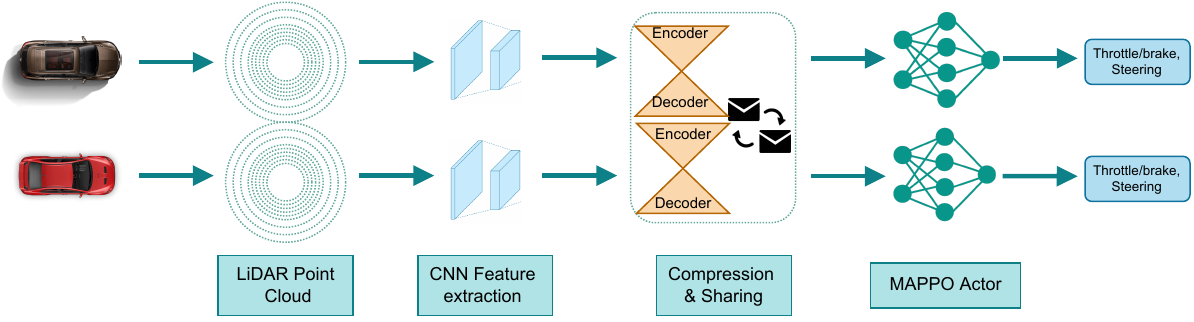}
  \caption{Collaborative MARL framework. LiDAR features are extracted from preprocessed point cloud data and shared within a V2V local communication network. The aggregated messages are then used to calculate the desired action through the actor network.}
  \label{fig:framework}
\end{figure*}

To ensure safe and efficient navigation through occluded intersections, we developed a collaborative framework based on the MAPPO algorithm and LiDAR features shared between vehicles using a V2V ad hoc network. In our proposed framework, each CAV is represented by an actor network that receives compressed LiDAR features along with metadata from nearby CAVs. A schematic representation of the proposed collaborative pipeline is presented in figure \ref{fig:framework}. 

The following is a description of the forward pass of the MARL collaborative mechanism used in a research article:

\begin{itemize}
    \item Step 1: Preprocessing and feature extraction. Individual LiDAR data is preprocessed and converted into ego vehicle coordinates. The processed data is passed through a series of convolutional layers to create meaningful features. 
    \item Step 2: Compression and V2V communication. The output features are compressed using the DRACO 3D compression algorithm \cite{noauthor_draco_2019} to reduce the message size by about 40 times. The agents can share the compressed features in a V2V ad hoc network. Each CAV then decompresses the message through a decoder, transforms the message to its perspective and aggregates the resulting features into a single tensor.
    \item Step 3: Action prediction. The MAPPO actor network pipeline is used to obtain the action. The input to the MAPPO actor network includes the aggregated LiDAR features.
\end{itemize}

The algorithm pseudocode is presented in \ref{alg:cap}. 

\begin{algorithm}
\caption{Collaborative MAPPO using LiDAR features.
}\label{alg:cap}
\begin{algorithmic}
\State Initialize $\theta$, the parameters for the critic $V$ and the parameters $\phi$ for the actor $\pi$.
\For {$ep=1$ to $M$}
\State Retrieve initial state $s_{0}, \Omega_{0}, \hat{z}_{i}$ 
\For {$t=1$ to $T$}

\For {$j$ in $vehicles$}
\State $h_i^{(0)} = CNN(\hat{z}_i^{(0)})$
\State $Compress(h_i^{(0)})$
\State $h_s = Aggregate(h_i)$
\State Choose action $a_{t}^{(j)} = \pi (\Omega_{t}^{(j)}, h_s; \theta)$
\State Calculate value $v^{(j)}=V(s_{t}^{(j)}, h_s; \phi)$
\EndFor 
\State Observe $r_{t}, s_{t+1}, \Omega_{t+1}, \hat{z}_i$.
\State Store $[s_{t}, \hat{z}_i, \Omega_{t}, a_{t}, r_{t}, s_{t+1}, o_{t+1}]$.

\State Compute advantage estimate $\hat{A}$ via GAE
\State Compute reward-to-go $\hat{R}$ on the stored transitions. 
\EndFor
\State Update $\theta$ using Adam optimiser on mini-batch
\State Update $\phi$ using Adam optimiser on mini-batch
\EndFor

\end{algorithmic}
\end{algorithm}

\section{EXPERIMENTS}

\begin{figure*}[t!]
  \centering
  \includegraphics[scale=0.6]{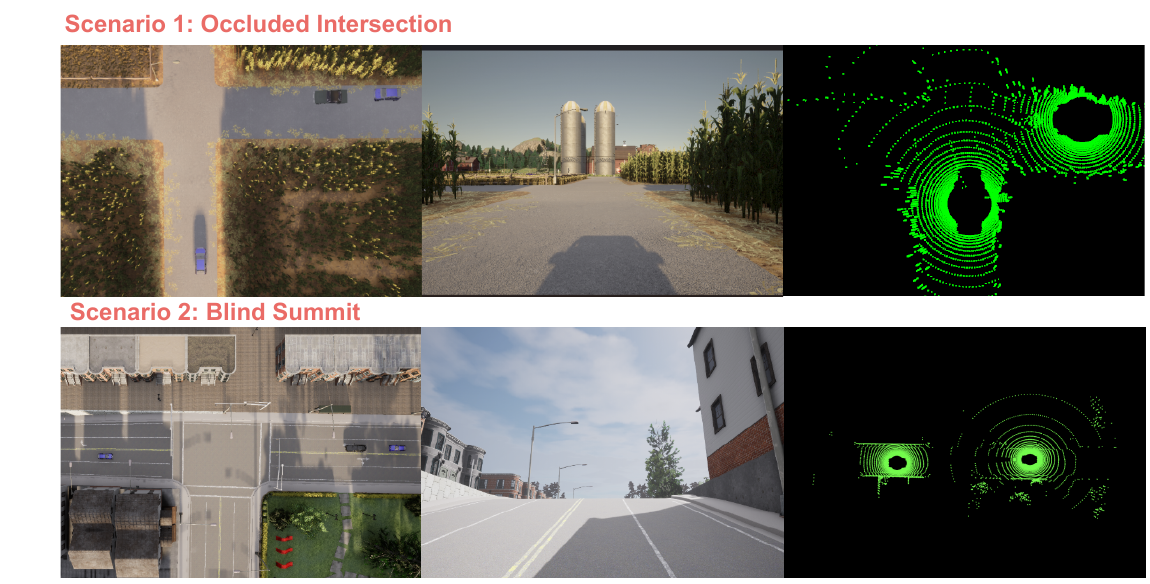}
  \caption{Occluded scenarios. The first scenario corresponds to a blind intersection, where vehicles are occluded by tall trees. The second scenario corresponds to a blind summit.}
  \label{fig:scenarios}
\end{figure*}

We design two different occluded intersection scenarios that are challenging for vehicles to navigate safely without causing any collisions. These scenarios consist of two CAVs and traffic vehicles. The latter are not part of the V2V network and, therefore, cannot share information (LiDAR data, position, velocity) with nearby CAVs. A Bird's Eye View of the scenarios is presented in figure \ref{fig:scenarios}. Scenario 1 consists of an intersection occluded by tall trees, while scenario 2 consists of a blind summit. Both scenarios are dangerous and consistently lead to collisions. The implementation details of the experiments can be found in the Appendix.

First, we compare our proposed collaborative MARL approach against the following baselines:

\begin{itemize}
    \item MAPPO with ground-truth locations: CAVs can observe the precise location of each vehicle in the scene, including the occluded vehicle/s. The rest of the pipeline structure is preserved. This baseline should outperform all other methods as it has full observability. The precise location of the vehicles may not be known in practice, or they may be a noisy estimate of the true locations.
    \item MAPPO with Early Fusion: CAVs share the raw LiDAR point cloud with other nearby agents and use it to predict the action. Sharing the entire LiDAR point cloud with nearby agents to predict actions is impractical due to bandwidth limitations in current communication protocols, such as such as DSRC and C-V2X \cite{cui_coopernaut_2022}.
    \item Independent RL: Each ego vehicle uses its own LiDAR observation and metadata to generate the control policy. Agents do not share information with each other and try to maximise an independent reward function. The MAPPO structure is the same for the multi-agent case, except for the message compression and sharing phase.
    \item Car following model: corresponds to a TTC-based model that calculates the safety time distance to the front vehicle and slows down in case it reaches a threshold. Otherwise, it tries to match the front vehicle's speed.
\end{itemize}

To facilitate a comprehensive comparison between the proposed approach and the baselines, we employ a set of commonly used metrics, consistent with those utilised in similar studies \cite{kamran_risk-aware_2020, isele_navigating_2018}:

\begin{itemize}
    \item Average Speed: Average speed per agent across all testing episodes. This is a measure of the overall traffic efficiency.
    \item Collision Rate: percentage of testing episodes that ended with a collision. This is an indicator of how safe the policies are generated by the different methods.
    \item Success Rate: The percentage of the test episodes where the agents reached the destination without collisions. A vehicle can successfully avoid collisions but may not reach the desired goal; therefore, the success rate reflects both safety and efficiency. 
\end{itemize}

\subsection{Baseline comparison}

Table \ref{table_baseline} presents a comparison of the proposed collaborative approach with different baselines. The results represent the average and standard deviation of 100 independent testing runs. The rule-based method has a high average speed but leads to collisions in almost a quarter of the episodes. The independent RL approach outperforms the rule-based method in terms of collision and success rate, but its average speed is relatively low. In contrast, the proposed collaborative method yields efficient driving policies with low collision rates.

\begin{table*}[h]
\centering
\caption{Comparison between the proposed collaborative method to the different baselines. Results correspond to the average over 1000 independent test runs.}
\label{table_baseline}
\resizebox{\textwidth}{!}{%
\begin{tabular}{@{}cccccccc@{}}
\toprule
Scenario & Method & Avg. Reward $\uparrow$&  Avg. Speed $(km/hr) \uparrow$ & Collision Rate $(\%)  \downarrow$ & Success Rate $(\%) \uparrow$  \\
 \bottomrule
 \multirow{4}{*}{Occluded Intersection}
  & Rule-based & $47.04 \pm 12.43$ & $40.45 \pm 5.62$ & $25.24 \pm 10.15$  & $62.44 \pm 10.23$  \\
  & Independent RL & $45.97 \pm 16.00$ & $40.65 \pm 15.45$ & $10.43 \pm 5.53$ & $76.66 \pm 17.96$ \\
  & MAPPO + Ground Truth & $68.23 \pm 6.06$ & $48.35 \pm 16.04$  &  $4.87 \pm 6.48$ & $ 89.12 \pm 12.77$  \\
  & MAPPO + Early Fusion & $59.03 \pm 5.32$ & $44.34 \pm 12.35$  &  $5.12 \pm 4.44$ & $ 78.45 \pm 9.45$  \\
  & Collaborative MAPPO (ours) & $63.35 \pm 8.05$ & $45.62 \pm 25.30$  &  $2.12 \pm 4.48$ & $ 85.34 \pm 15.13$  \\
  \hline  
   \multirow{4}{*}{Blind Summit}
  & Rule-based & $210.25 \pm 12.12$ & $60.14\pm 8.16$ & $17.45 \pm 10.14$  & $68.24 \pm 18.12$  \\
  & Independent RL & $180.45 \pm 44.71$ & $68.42 \pm 16.72$ & $12.15 \pm 8.86$ & $70.71 \pm 17.14$ \\
  & MAPPO + Ground Truth & $270.71 \pm 23.25$ & $75.04 \pm 14.99$  &  $1.91 \pm 3.32$ & $ 96.94 \pm 10.14$  \\
  & MAPPO + Early Fusion & $235.77 \pm 31.75$ & $71.12 \pm 21.22$  &  $4.24 \pm 2.22$ & $ 90.90 \pm 5.28$  \\
  & Collaborative MAPPO (ours) & $257.24 \pm 36.76$ & $71.08 \pm 20.98$  &  $3.07 \pm 3.15$ & $94.14 \pm 7.15 $  \\

 \bottomrule
\end{tabular}}
\end{table*}



The collaborative framework demonstrates comparable performance to using ground truth locations of all vehicles in the scene. Although the collision rates are similar, the $MAPPO + Ground Truth$ method achieves higher traffic efficiency and, therefore, higher overall reward. This can be attributed to the fact that the true locations of all vehicles are part of the observation, while the proposed method relies on a fixed range for transmitting and receiving LiDAR features. Thus, agents become aware of occluded vehicles only when other CAVs are within communication range. In real-world situations, having precise vehicle locations may not be feasible, especially when vehicles are occluded by buildings or other obstacles.

\begin{figure}[t!]
  \centering
    \includegraphics[scale=0.4]{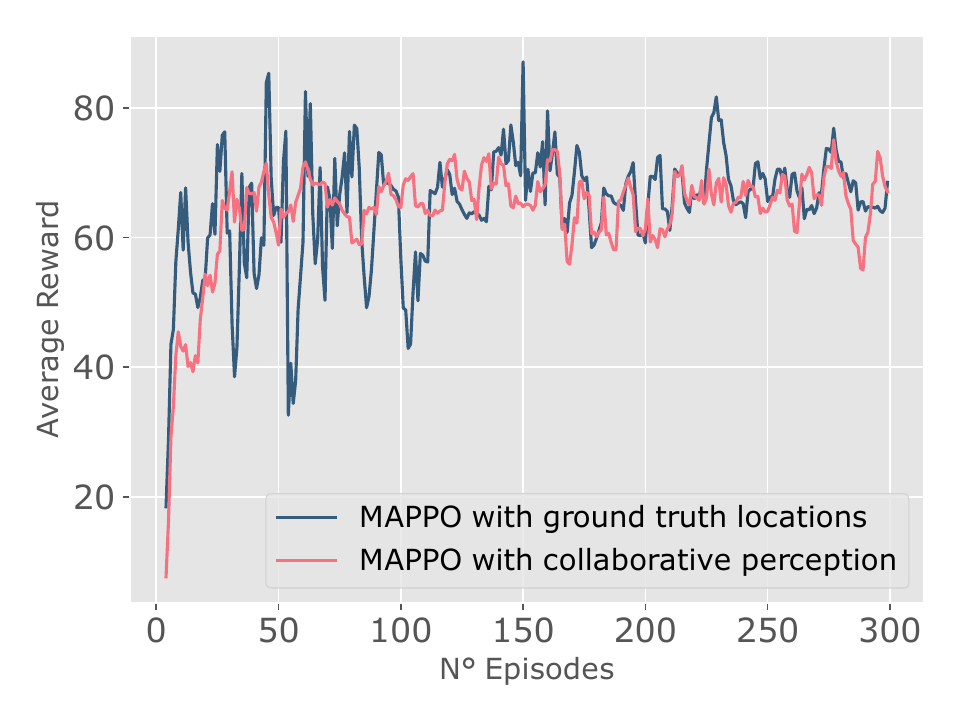}
  \caption{Learning curves for MAPPO with ground truth locations of occluded vehicles and MAPPO with collaborative perception.}
  \label{fig:ground_truth_vs_collab}
\end{figure}

Figure \ref{fig:ground_truth_vs_collab} compares the learning curves of MAPPO trained with ground truth locations of all vehicles and MAPPO with the collaborative perception pipeline. The results, which are an average of 5 independent training runs, show that both methods achieved a similar final average reward. However, the proposed method had a more stable training phase, particularly during the first 100 episodes, which could be attributed to the additional information contained in LiDAR features that aided in stabilising the learning process.

The proposed method effectively meets the bandwidth requirements of current communication protocols such as DSRC and C-V2X. With a frame rate of 20 fps and convolutional features represented as (4, 128, 128), the bandwidth utilisation, after employing the Draco compression algorithm, amounts to approximately 1.075 Mbps. This value remains comfortably below the throughput capabilities of V2X communication protocols, with DSRC at 2.0 Mbps and C-V2X at 7.2 Mbps \cite{cui_coopernaut_2022}. It is worth noting that transmitting the complete point cloud, which consists of 47,240 points, requires approximately 80 Mbps, which exceeds the allowable bandwidth limits.



\subsection{Point Dropout resilience}

Real-world liDAR data can suffer from high levels of noise and point dropout \cite{espadinha_lidar_2021}. This creates a gap between simulation and reality; thus, control methods should be robust to different noise levels in the sensor reading. To test the robustness of the proposed framework, we simulate different levels of point dropout on the sensor data readings. We tested the trained algorithm with dropout levels ranging from $10\%$ to $40\%$ on the Occluded Intersection scenario.

\begin{figure}[h]
  \centering
  \includegraphics[scale=0.4]{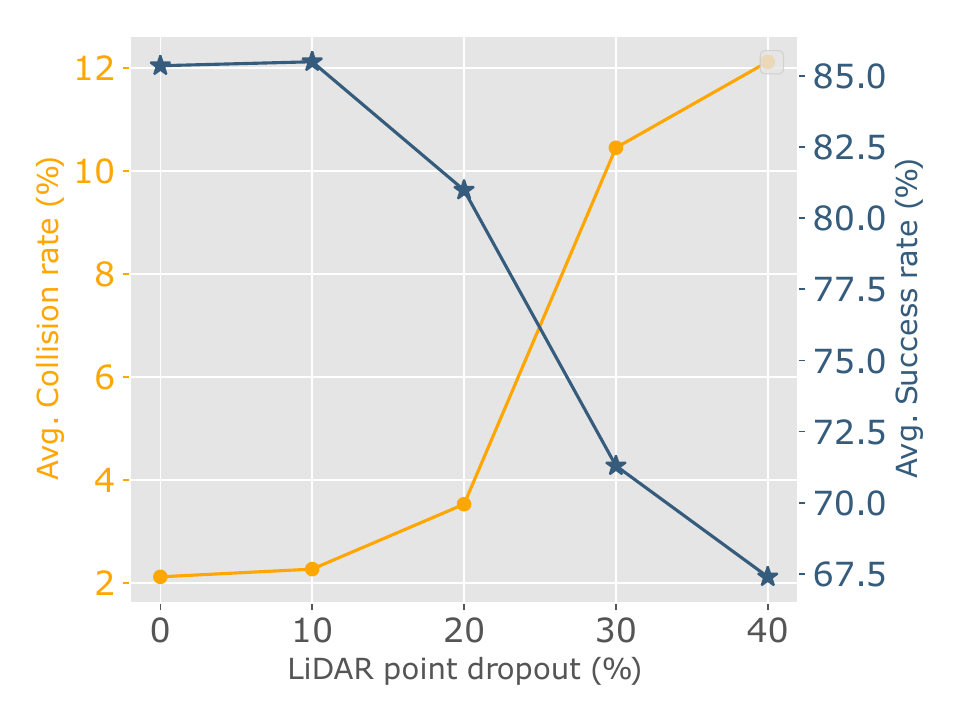}
  \caption{Collision and Success rate as a function of LiDAR point dropout.}
  \label{fig:dropout}
\end{figure}

The proposed method can still perform well in scenarios with $20\%$ LiDAR point dropout or less. Figure \ref{fig:dropout} shows the average test collision rate of the collaborative approach under different levels of point dropout. $20-30\%$ dropout represents the threshold region where performance significantly degrades, with collision rates increasing five-fold and, consequently, underperforming compared to independent RL.

\section{CONCLUSIONS}

Collaborative control methods become essential for occluded scenarios where one or more vehicles are blocked, for instance, by another car or tall building. We presented a decentralised approach based on a local V2V sharing network that can generate cooperative policies in challenging driving settings. The method aggregates LiDAR features of nearby CAVs and uses them to determine the action that maximises traffic efficiency while minimising collision risk. We compared the proposed method against a series of baselines, including a car following baseline, independent RL and a cooperative early fusion method.

Generating collaborative policies leads to lower levels of collisions and higher traffic efficiency. Utilising the proposed method with LiDAR feature sharing, the rate of crashes can be reduced on average by $70\%$ compared to independently trained policies. It also leads to more stable learning, suggesting that using LiDAR features can help stabilise the training procedure. Moreover, $20\%$ of point cloud dropout does not have a great effect on algorithm performance. With higher dropout levels, the performance degrades substantially, leading to underperforming compared to independent RL policies.

Even though the proposed method substantially reduces the risk of collision in occluded scenarios, some collisions are still present. Thus, other methods should also be studied that can guarantee no collisions, likely leading to lower traffic efficiency.




\section*{APPENDIX}

\subsection{Implementation details}

The parameters for the gym environment are presented in table \ref{parameters_env}. We consider a maximum communication range of 70m, which is realistic given current autonomous driving communication protocols, such as DSRC \cite{kenney_dedicated_2011}. All experiments were conducted using a 10-core, 3.70Ghz Intel Core i9-10900X CPU and an Nvidia GTX 3090 GPU. 200 steps were simulated for the Occluded intersection, while 400 were carried out for the Blind Summit, because the latter has a larger map.

\begin{table}[h]
\footnotesize
\centering
\caption{Environment parameters.}
\label{parameters_env}
\begin{tabular}{@{}cc@{}}
\toprule
Parameter   & Values 
\\ \toprule
Vehicle Min. Speed & $0 km/hr$ \\
Vehicle Max. Speed &  $80 km/hr$   \\
Communication range & $70 m$\\
N° LiDAR Channels & $32$ \\
LiDAR Range & $50m$ \\
LiDAR points per second & $1^6 points/s$\\

\bottomrule
\end{tabular}
\end{table}




\bibliographystyle{IEEEtran} 
\bibliography{IEEEabrv,marl}

\end{document}